\documentclass[conference]{IEEEtran}

\usepackage[utf8]{inputenc}
\usepackage[T1]{fontenc}
\usepackage{cite}
\usepackage{amsmath,amssymb,amsfonts}
\usepackage{graphicx}
\usepackage{booktabs}
\usepackage{array}
\usepackage[table]{xcolor}
\usepackage{balance}
\usepackage{tikz}
\usepackage{siunitx}
\usetikzlibrary{arrows.meta,positioning,fit,backgrounds,calc}
\newcommand{\tph}{--}
\providecommand{\Description}[1]{}

\newcommand{\localmodel}{\textsc{Local}}
\newcommand{\fmzero}{\textsc{Zero-Shot}}
\newcommand{\fmfine}{\textsc{Fine-Tuned}}
\newcommand{\fmpool}{\textsc{Pooled}}
\newcommand{\cvmodel}{\textsc{Const.-Vel.}}
\newcommand{\idmmodel}{\textsc{IDM}}

\begin{document}

\title{PhaseShift: Topology-Aware Data Harmonization and Model Consolidation Across Signalized Intersections}

\author{\IEEEauthorblockN{Yash Ranjan, Artur Kumik, Rahul Sengupta, Anand Rangarajan, Sanjay Ranka}
\IEEEauthorblockA{University of Florida, Gainesville, FL, USA}}

\maketitle

\begin{abstract}
Learned traffic-behavior models are commonly trained separately for each intersection, creating model portfolios that cannot share evidence across sites. We present \textbf{PhaseShift}, a topology-aware framework that harmonizes heterogeneous roadside trajectories into a shared actor-centric representation and trains one reusable backbone. Ego-relative coordinates, trajectory-induced movement paths, normalized signal context, and variable-cardinality interaction tokens remove site conventions while preserving behaviorally relevant topology. The backbone supports pooled operation, zero-shot use at a held-out intersection, and low-data adaptation. We evaluate five intersections in two Florida regions on balanced field data---100{,}000 training windows and equal-sized test sets per site---under a replay-conditioned, best-of-sampled-trajectory protocol. At 10~s, one pooled model lowers both minADE and minFDE relative to independently trained local models at all five sites, with median reductions of 36.8\% and 22.0\%. Leave-one-intersection-out deployment, including one cross-region fold, beats local training on both 10-s metrics at four of five sites, although short-horizon performance is less uniform. Fine-tuning with 1{,}000 target update windows improves on zero-shot at three sites and is the strongest regime at one. At site~7, every cross-site mixture sharply lowers long-horizon error under a fixed 100{,}000-window budget; test-likelihood gains argue against a best-of-sample dispersion-only explanation. Local models fall behind calibrated IDM at the two highest-flow sites after long autoregressive rollouts; pretrained-backbone regimes do not. Within this five-site evaluation, PhaseShift demonstrates consolidation across heterogeneous physical--control settings while identifying sites that still require adaptation. The protocol measures conditional single-vehicle generation under replayed context, not closed-loop traffic simulation.
\end{abstract}


\section{Introduction}
Deploying learned traffic models across a road network can create a portfolio of narrowly specialized models. If every intersection and operating condition is treated as an independent dataset, five intersections observed across seven day-of-week domains can require as many as 35 models. Training, calibrating, validating, and maintaining each model separately is costly, and observations collected in one domain cannot strengthen another. The point is not that every recurring condition warrants a separate model; it is that a site-specific workflow provides no principled reuse path when a new local condition appears.

This paper asks whether heterogeneous intersection data can instead support one shared model without erasing the local context that governs behavior. Our workflow harmonizes the data, learns a reusable backbone, deploys it directly where it generalizes, and adapts only the domains with a meaningful residual. It follows the pretraining-and-reuse pattern, but our claims are limited to regional harmonization and consolidation. We use \emph{site--condition domain} for an intersection under a recurring operating condition, such as a time period, signal plan, or demand regime; the present study treats each intersection as one domain.

Signalized intersections make this difficult because observed trajectories arise from an entangled physical system. Topology determines feasible movements and conflicts; signal control determines when movements are released; demand shapes queues and interaction frequency; and road users respond heterogeneously. Roadside sensing adds distinct coordinate frames and controller conventions, incomplete tracks, occlusion, and tracking noise. Cross-site learning therefore requires removing incidental conventions while retaining the physical and control context that generates behavior.

Classical microscopic simulators combine authored road networks with explicit car-following, lane-changing, gap-acceptance, and signal-control rules~\cite{NI2020102137}. Learned motion models estimate behavior from trajectories, but prior intersection systems are commonly developed around one or a small number of authored geometries~\cite{metrics,ranjan2025,enactor2026}. In roadside deployments, geometry, control, demand, human behavior, and sensing quality vary jointly rather than as independently controlled factors. Cross-site modeling is therefore a \emph{physical-AI} problem: learning reusable structure from noisy measurements of changing physical--control regimes.

We introduce \textbf{PhaseShift}, a topology-aware framework for multi-intersection data harmonization and model consolidation. PhaseShift converts each roadside scene into a common actor-centric representation using ego-relative coordinates, trajectory-induced movement paths, normalized signal state, stop lines, and variable-cardinality road-user tokens. One shared actor-conditional backbone is pretrained across multiple intersection domains without a learned site identifier, then used as one pooled model over known domains, zero-shot at a held-out intersection, or as initialization for limited target adaptation.

PhaseShift makes three contributions:
\begin{itemize}
    \item \textbf{Topology- and control-aware data harmonization.} We map trajectories, neighboring actors, pedestrians, movement geometry, stop lines, and signal states into a shared actor-centric token space while retaining the local physical--control structure needed for conditional trajectory generation.
    \item \textbf{Shared pretraining and specialization.} One backbone supports pooled operation, leave-one-intersection-out zero-shot deployment, and low-data adaptation. Per-domain residuals distinguish sites covered by the common model from those that warrant specialization or investigation.
    \item \textbf{A five-domain field evaluation.} On balanced trajectories from five intersections in two Florida regions, we compare one pooled model with independently trained local models, evaluate every held-out site, adapt every target, and run a fixed-window-count composition sweep at one site as a first control on the diversity--volume confound.
\end{itemize}

\section{Related Work}
\label{sec:related}

\paragraph*{Reusable pretraining and foundation-model workflows}
The foundation-model paradigm pretrains reusable representations over heterogeneous data and adapts them downstream rather than training every model from scratch. Transportation work has explored graph pretraining~\cite{wang2023buildingtransportationfoundationmodel}, multi-scene motion models~\cite{trafficbots2023,smart2024}, and city-scale world models~\cite{tan2025scenediffusercityscaletrafficsimulation}. PhaseShift targets the regional deployment problem: harmonizing field trajectories from topologically different intersections into one model for pooled use, held-out deployment, and selective adaptation.

\paragraph*{Microscopic traffic simulation}
Rule-based simulators such as SUMO and MATSim combine authored road networks with car-following, lane-changing, gap-acceptance, route-choice, and signal-control models~\cite{NI2020102137}. They remain indispensable for traffic operations, but behavior rules and parameter distributions must be selected and calibrated for each application. At intersections, aggregate calibration can reproduce flow or delay while missing heterogeneous turning, yielding, and queue-discharge behavior. PhaseShift is complementary to a broader simulator: the present paper evaluates a reusable conditional vehicle-trajectory model under replayed scene context, not a reciprocal closed-loop traffic simulator.

\paragraph*{Learned motion and intersection behavior}
Trajectory-prediction and learned world-model systems use multi-agent context, vectorized maps, and multimodal outputs to generate future motion~\cite{trajsurvey1,trajsurvey2,salzmann2021,shi2023mtr,hdgt2023,gao2020}, and prior intersection models incorporate signal state, movement intent, and interaction context for long-horizon generation and, in some systems, scene simulation~\cite{intersectionnet,kigan2024,vehits23,metrics,ranjan2025,enactor2026}. These methods establish important architectural components, but they do not by themselves solve the dataset-harmonization problem created by roadside collections with distinct coordinates, movement vocabularies, and controller conventions.

\paragraph*{Roadside sensing and cross-domain deployment}
Video and LiDAR systems can recover vehicle and pedestrian trajectories, signal state, and conflict measures directly from instrumented intersections~\cite{BanerjeeCASKGPP22,mishra2022using}. Such data differ from curated vehicle-centric benchmarks: the observed region is fixed, tracks can start or end inside the field of view, and each recording reflects one realized joint physical--control regime. The model must bridge measurements from these regimes rather than rely on one normalized synthetic environment. PhaseShift treats local geometry, signal state, and observation validity as explicit context rather than learned site identities.

\section{Multi-Intersection Pretraining Formulation}
\label{sec:method}
Let $\mathcal{D}=\{D_1,\ldots,D_M\}$ denote a collection of site--condition domains. A domain may correspond to an intersection, or more finely to an intersection under a recurring condition. Each domain represents a joint physical--control regime rather than topology alone. Because raw scenes also use different coordinates, movement paths, signal identifiers, and visible actor sets, PhaseShift applies a domain-specific but non-learned harmonization operator $h_m$ that converts each scene into a common actor-centric context using local geometry and normalized signal state.

The shared model parameters are pretrained over the harmonized domains:
\begin{equation}
    \theta^\star = \arg\min_{\theta}
    \sum_{m=1}^{M} w_m\,
    \mathbb{E}_{(S,Y)\sim D_m}
    \left[\mathcal{L}\!\left(f_\theta(h_m(S)),Y\right)\right],
    \label{eq:pretraining}
\end{equation}
where $w_m$ controls domain sampling and $f_\theta$ is the shared actor-conditional model. No learned site identifier is required. For known domains, the same $\theta^\star$ is the pooled operational model. For an unseen target $q$, $h_q$ supplies its geometry and signal context while $\theta^\star$ is used zero-shot; with a small target set, the model is initialized from $\theta^\star$ and fine-tuned.

Let $S^t$ denote the recorded scene at time $t$, including actor states, pedestrian observations, induced movement geometry, stop lines, and normalized signal state. For a target actor $i$, PhaseShift learns the shared actor-conditional transition model
\begin{equation}
    p_\theta\!\left(s_i^{t+1}\mid \mathcal{C}_i^{0:t}\right),
\end{equation}
where $s_i^{t+1}$ is the next state of actor $i$ and $\mathcal{C}_i^{0:t}=h_m(S^{0:t};i)$ is the harmonized history expressed in actor $i$'s local frame. The context contains the target history, currently valid neighboring observations, local movement geometry, stop lines, and signal state; demand appears through evolving actor and movement occupancy rather than a learned site label.

The evaluated $H$-step trajectory distribution is generated autoregressively,
\begin{equation}
    p_\theta\!\left(s_i^{t+1:t+H}\mid \mathcal{C}_i^{0:t}\right)
    = \prod_{\tau=1}^{H}
    p_\theta\!\left(s_i^{t+\tau}\mid \mathcal{C}_i^{0:t+\tau-1}\right),
    \label{eq:actorrollout}
\end{equation}
with the generated target state fed back after every step; neighboring actors, pedestrians, and signal state are replayed from the logged scene.

This formulation separates three concerns: how a site's raw observations are expressed in the common token space (the harmonizer), behavior learned across the regional collection (the shared parameters), and residual behavior the common model does not capture (optional fine-tuning).

\section{Cross-Domain Data Harmonization}
\label{sec:repr}
The harmonization operator converts site-native observations into a shared token space while preserving the geometry, control state, and actor context that determine behavior. It removes incidental conventions---coordinate origin, orientation, controller identifiers, and fixed actor ordering---without normalizing away structured variation in demand, phasing, or human response. Every entity is represented by a vector token with ego-relative pose and a type-specific descriptor of dynamics or extent.

\paragraph*{Ego-relative pose}
For ego position $(x_i,y_i)$ and heading $\psi_i$, all positions and headings are translated and rotated into the ego frame. An entity is represented by $(\Delta x,\Delta y,\Delta\psi)$, with angular quantities encoded continuously where appropriate. Translating or rotating the entire intersection thus leaves the tokenized scene unchanged (\emph{global-pose invariance}); this removes site orientation and coordinate origin without claiming invariance to the local geometry itself.

\paragraph*{Factorized token encoding}
Pose and descriptor are encoded separately before being combined. The descriptors are:
\begin{itemize}
    \item \textbf{Neighbor actors:} actor type, speed, heading change, and any additional valid kinematic features, combined with relative pose.
    \item \textbf{Lane geometry:} short directed segments sampled from representative movement polylines, with segment length and orientation represented in the ego frame.
    \item \textbf{Stop lines and signals:} stop-line segments with the associated normalized phase state attached to the governed geometric element.
\end{itemize}
The model receives no intersection identifier, absolute coordinate, or learned site embedding.

\paragraph*{Signal and conflict context}
Controller-specific phase identifiers are converted to a site-independent categorical representation attached to the affected stop line or movement, such as red, yellow, protected green, permissive green, or unknown. This prevents the model from memorizing site-specific controller labels while preserving operational state. Pedestrians are encoded as dynamic contextual actors---so a turning vehicle can condition on crosswalk occupancy---though they are not generated in the current experiments.

\paragraph*{Trajectory-induced geometry}
PhaseShift replaces an authored lane-connectivity graph with compact movement polylines induced from trajectories collected at and around each intersection. The geometry-building interval is disjoint from behavioral training, validation, and test data. The two components have different data requirements: geometry is constructed only from tracks complete enough to identify an approach and departure, whereas the behavioral model consumes arbitrary partial tracks at inference time.

The procedure has four steps. (i)~\emph{Movement grouping:} sufficiently complete vehicle tracks are assigned to a movement class by their starting and ending bounding boxes (entry and exit approaches); turn type is inferred from the path through the intersection. (ii)~\emph{Arc-length normalization:} tracks within a movement are resampled to common arc-length positions for geometric comparability across speeds. (iii)~\emph{Summarization:} a single representative path per movement is computed by aligning the cluster's trajectories with dynamic time warping and is discretized into short directed segments consumed by the encoder. (iv)~\emph{Support filtering:} movement classes below a minimum support threshold are excluded from the induced geometry rather than represented by an unreliable path. The resulting site descriptor captures the approaches, turn paths, and conflict-zone traversal observed in roadside data without requiring a manually authored lane graph or target-site behavior labels.

The principal experiments hold this geometry-building procedure fixed while evaluating behavior transfer.

\fmzero{} is therefore zero-shot with respect to target behavior-model parameter optimization, not zero-data deployment: the target still supplies movement geometry, stop lines, and observed signal context. The geometry-building interval remains disjoint from behavioral evaluation; its per-site data requirement is part of the deployment cost that the present study does not quantify.

\paragraph*{Incomplete observations}
At each time step the spatial encoder receives only the entities marked valid by the tracker, with no requirement that the same actors be present at every history step; a missing observation removes a token rather than imputing a dense fixed-size agent tensor. This variable-cardinality design accommodates the incomplete and changing observations produced by roadside tracking without asserting robustness beyond the reported field evaluation.

\section{Shared Backbone and Deployment Lifecycle}
\label{sec:arch}
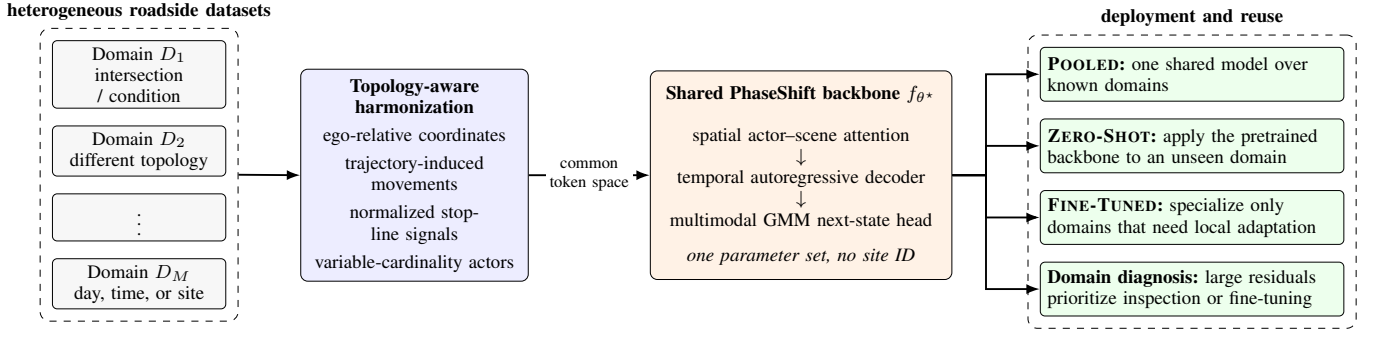
\begin{figure*}[t]
    \centering
    \resizebox{0.99\textwidth}{!}{%
    \begin{tikzpicture}[
        font=\footnotesize,
        dombox/.style={draw, rounded corners=2pt, align=center, minimum height=7mm, text width=24mm, inner sep=2.5pt, fill=black!3},
        harmonize/.style={draw, rounded corners=3pt, align=center, minimum height=27mm, text width=31mm, inner sep=4pt, fill=blue!7},
        backbone/.style={draw, rounded corners=3pt, align=center, minimum height=31mm, text width=42mm, inner sep=4pt, fill=orange!10},
        deploy/.style={draw, rounded corners=2pt, align=left, minimum height=8mm, text width=43mm, inner sep=3pt, fill=green!7},
        stage/.style={draw, dashed, rounded corners=4pt, inner sep=5pt},
        ar/.style={-{Latex[length=2mm]}, thick},
    ]
    \node[dombox] (d1) {Domain $D_1$\\intersection / condition};
    \node[dombox, below=2.5mm of d1] (d2) {Domain $D_2$\\different topology};
    \node[dombox, below=2.5mm of d2] (d3) {$\vdots$};
    \node[dombox, below=2.5mm of d3] (dm) {Domain $D_M$\\day, time, or site};
    \begin{scope}[on background layer]
        \node[stage, fit=(d1)(dm), label={[font=\footnotesize\bfseries]north:heterogeneous roadside datasets}] (domains) {};
    \end{scope}

    \node [harmonize, right=11mm of d2.south east, yshift=0.2mm] (harm) {\textbf{Topology-aware harmonization}\\[1mm]
    ego-relative coordinates\\[1mm]
    trajectory-induced movements\\[1mm]
    normalized stop-line signals\\[1mm]
    variable-cardinality actors};

    \node[backbone, right=18mm of harm] (fm) {\textbf{Shared PhaseShift backbone $f_{\theta^\star}$}\\[3mm]
    spatial actor--scene attention\\$\downarrow$\\temporal autoregressive decoder\\$\downarrow$\\multimodal GMM next-state head\\[2mm]
    \emph{one parameter set, no site ID}};

    \node[deploy, right=13mm of fm, yshift=15mm] (pool) {\textbf{\fmpool:} one shared model over known domains};
    \node[deploy, below=2.5mm of pool] (zs) {\textbf{\fmzero:} apply the pretrained backbone to an unseen domain};
    \node[deploy, below=2.5mm of zs] (ft) {\textbf{\fmfine:} specialize only domains that need local adaptation};
    \node[deploy, below=2.5mm of ft] (diag) {\textbf{Domain diagnosis:} large residuals prioritize inspection or fine-tuning};
    \begin{scope}[on background layer]
        \node[stage, fit=(pool)(diag), label={[font=\footnotesize\bfseries]north:deployment and reuse}] (uses) {};
    \end{scope}

    \draw[ar] (domains.east) -- (harm.west);
    \draw[ar] (harm.east) -- node[midway, fill=white, align=center, font=\scriptsize, inner sep=1.5pt] {common\\token space} (fm.west);
    \draw[ar] (fm.east) -- ++(5mm,0) |- (pool.west);
    \draw[ar] (fm.east) -- ++(5mm,0) |- (zs.west);
    \draw[ar] (fm.east) -- ++(5mm,0) |- (ft.west);
    \draw[ar] (fm.east) -- ++(5mm,0) |- (diag.west);
    \end{tikzpicture}%
    }
    \caption{PhaseShift's topology-aware multi-intersection pretraining and deployment workflow. A shared spatial--temporal backbone is trained across harmonized domains and reused as one pooled model, deployed zero-shot at a new intersection, or fine-tuned where local data show that specialization is needed. The domain-diagnosis path treats residual zero-shot error as a prioritization signal, not as proof of a particular behavioral cause.}
    \Description{A four-stage diagram shows heterogeneous roadside domains flowing through topology-aware harmonization into one shared PhaseShift backbone. The backbone supports pooled use, zero-shot deployment, selective fine-tuning, and identification of domains requiring additional inspection.}
    \label{fig:arch}
\end{figure*}

PhaseShift processes each harmonized scene in two stages: a per-timestep spatial encoder aggregates the variable-cardinality scene into an ego embedding, and an autoregressive temporal decoder predicts a multimodal next-state distribution. Figure~\ref{fig:arch} places this backbone inside the full multi-intersection pretraining-and-reuse lifecycle.

\subsection{Stage 1 --- Spatial encoding}
Stage~1 runs per timestep with weights shared across the sequence.

\paragraph*{Tokenizers}
Each modality is projected to a common $d_{\text{model}}=256$ space by lightweight $\text{Linear}\!\to\!\text{LeakyReLU}\!\to\!\text{Dropout}$ blocks. Actor-state and map-speed tokenizers are shared across contexts, while the relative-pose encoders are type-specific.

\paragraph*{Additive composition}
Each token is formed additively,
\begin{equation}
\text{token} = \underbrace{\phi_{\text{kin}}(\cdot)}_{\text{own kinematics}} + \underbrace{\phi_{\text{pose}}(\Delta x, \Delta y, \Delta \psi)}_{\text{relative pose}},
\end{equation}
with an additional $\phi_{\text{sig}}(\cdot)$ term for signal tokens---a factorized encoding that separates ``what an agent is doing'' from ``where it is relative to me.''

\paragraph*{Neighbor cross-attention}
A \texttt{Neighbor\-Attention\-Layer} takes the ego embedding as the query and all scene tokens (neighbors, lane polylines, and signal) as keys and values. Stacked multi-head attention layers, each with a residual connection and layer normalization, aggregate the whole scene into a single ego embedding per timestep. The output is a sequence of scene-aware ego embeddings of shape $[\text{batch}, \text{seq\_len}, d_{\text{model}}]$.

\subsection{Stage 2 --- Temporal decoding}
Stage~2 is an autoregressive attention decoder. At each future step, the most recent embedding is the query and the history embeddings are keys and values (stacked cross-attention layers).

\paragraph*{Relative position encoding (RPE)}
A learned pairwise embedding $\text{RPE}^{\,t,t-\tau}$ encodes each history state's pose relative to the current query, $(\Delta x,\Delta y,\sin\Delta\theta,\cos\Delta\theta,\Delta t)$, and is added to the keys and values. During training these relative poses are obtained from the teacher-forced trajectory; during rollout they are recomputed from generated states. This injects spatial displacement and elapsed time into temporal attention without exposing a global coordinate frame.

\paragraph*{Teacher forcing}
During training, the decoder uses teacher forcing: the next query is constructed from the ground-truth state history. At inference, the sampled next state is re-encoded and appended to the sliding history window.

\paragraph*{Probabilistic output head}
A Gaussian Mixture Model (GMM) head with $K=25$ modes uses separate output layers for the mixture weights. Each mode predicts a mean $(\mu_x,\mu_y)$ and diagonal log-scales $(\log\sigma_x,\log\sigma_y)$. The mixture captures multimodal uncertainty (for example, turn versus go straight).

\subsection{Training objective}
Let $m_i^k\in\{0,1\}$ indicate whether actor $i$ has a valid target at future step $k$. The model minimizes the masked negative log-likelihood of the ground-truth displacement under the predicted Gaussian mixture:
\begin{equation}
\mathcal{L}_{\mathrm{NLL}} =
-\frac{1}{M}
\sum_{i,k} m_i^k
\log\!\left[
\sum_{j=1}^{K}\pi_{i,j}^{k}\,
\mathcal{N}\!\big(
\Delta\mathbf{p}_{i,\mathrm{gt}}^{k}\mid
\boldsymbol{\mu}_{i,j}^{k},
\boldsymbol{\Sigma}_{i,j}^{k}
\big)\right],
\end{equation}
where $M=\sum_{i,k}m_i^k$ counts valid targets. Here $\boldsymbol{\Sigma}_{i,j}^{k}$ is diagonal and parameterized through predicted log standard deviations. Masking permits partial tracks to contribute wherever a valid target exists without discarding the surrounding scene.

\section{Data and Multi-Intersection Protocol}
\label{sec:data}
We use roadside trajectory and signal data from five signalized intersections in two Florida regions. Four sites are in Gainesville: intersection~$7$ (NW 13th St \& University Ave), $8$ (NW 17th St \& University Ave), $9$ (Gale Lemerand Dr \& University Ave), and $10$ (NW 23rd Ave \& NW 55th St). Intersection~$22$ (Stirling Rd \& SR-7) is a high-volume South Florida arterial intersection and supplies the cross-region target.

Each recording covers the conflict zone and surrounding approaches, and each intersection is treated as one pretraining domain. A fisheye camera records at 10 frames per second; a YOLO-based detector with DeepSORT tracking produces road-user tracks, and a thin-plate-spline transformation maps them to a rectangular ground frame. The intersection-level domain is the coarsest useful partition; the same site can later be divided by day, time, weather, signal plan, or demand regime.

Table~\ref{tab:sites} summarizes the five domains. They differ widely in the span of traffic behind the balanced training sets (19--84 minutes), demand (roughly 985--5062 observed vehicles per hour), movement count (8--24), and signal operation (cycle lengths of 117--179~s with different green splits and cycle variability), while the four Gainesville sites additionally share regional driving norms that site~22 does not. Dividing each training span by the nominal cycle length, the training data cover only about 19, 22, 28, 43, and 6 signal cycles respectively---a reminder that each domain captures few independent realizations of the signal-controlled dynamics, most severely at sites~7 and 22. Vehicles are generated by the model, while pedestrians are dynamic context in the current experiments.

\begin{table}[t]
\centering
\caption{Characteristics of the five intersection domains. ``Train'' is the duration of recorded traffic from which each site's 100{,}000 training samples are drawn; the total collected recordings are longer (0.42--1.75~h per site). Flow is the rate of vehicles assigned to a movement cluster (clustering uses each track's starting and ending bounding boxes; vehicles missing either endpoint are excluded); Movements is the number of induced movement clusters; CoV is the coefficient of variation of the signal cycle length; green split is the mean green fraction.}
\label{tab:sites}
%\scriptsize
\setlength{\tabcolsep}{2.6pt}
\begin{tabular}{crrrrrr}
\toprule
\textbf{Site} & \shortstack{\textbf{Train}\\\textbf{(min)}} & \shortstack{\textbf{Flow}\\\textbf{(veh/h)}} & \textbf{Movements} & \shortstack{\textbf{Cycle}\\\textbf{(s)}} & \shortstack{\textbf{Cycle}\\\textbf{CoV}} & \shortstack{\textbf{Mean green}\\\textbf{split (\%)}} \\
\midrule
7  & 48 & 2672.93 & 16 & 150.0 & 0.1 & 19.6 \\
8  & 48 & 1126.07 & 11 & 131.0 & 0.3 & 27.8 \\
9  & 56 &  985.37 &  8 & 120.0 & 0.3 & 46.0 \\
10 & 84 & 1332.15 & 14 & 117.0 & 0.3 & 27.6 \\
22 & 19 & 5061.54 & 24 & 179.0 & 0.0 & 24.0 \\
\bottomrule
\end{tabular}
\end{table}

\begin{figure}[t]
\centering
\includegraphics[width=\columnwidth]{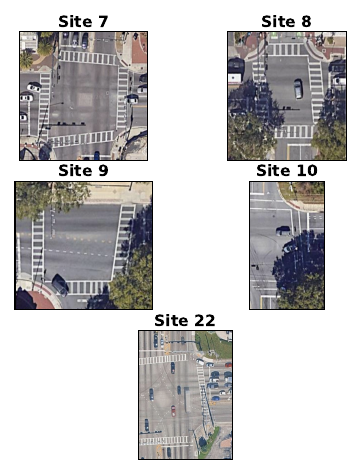}
\caption{The five intersection domains used by PhaseShift. The panels are cropped around the monitored conflict zones and are not shown at a common scale.}
\Description{Five aerial intersection images show sites 7, 8, 9, 10, and 22 with visibly different approach widths, lane arrangements, crosswalks, and roadside context.}
\label{fig:sites}
\end{figure}

Figure~\ref{fig:sites} shows the aerial layouts for all five intersections. The domains differ in footprint, approach width, lane arrangement, turn movements, crosswalk placement, and surrounding occlusion, illustrating why a common coordinate frame or fixed lane vocabulary is insufficient. Each recording also samples one realized combination of signal control, directional demand, human response, and sensing conditions; it is therefore a physical--control domain, not a topology-only dataset. The views are illustrative rather than metrically aligned.

\paragraph*{Balanced sampling}
Because the recordings differ in duration and demand, we balance the datasets rather than use them raw. Each intersection contributes 100{,}000 training samples; one sample contains a 20-step (2-s) observed history and a 20-step future supervision window for one target vehicle. Under a per-intersection seed (a site hash combined with one global seed), each site's window pool is permuted once and the first 100{,}000 windows are selected without replacement. Smaller fractions are prefixes of this permutation and therefore nested within larger draws, making the composition sweep directly comparable. Evaluation likewise uses equal-sized held-out test sets, so no site dominates by volume. The eligible-window populations at sites 7/8/9/10/22 are 282{,}411/126{,}667/110{,}644/106{,}451/142{,}081, and the draw consumes 35--94\% of each pool. The analysis does not count unique vehicles; because samples are overlapping windows from continuous tracks, nominal counts overstate effective sample size.

\paragraph*{Partitioning}
Each recording is partitioned chronologically into disjoint movement-geometry, behavioral-training, validation, and held-out test intervals. Validation uses 12{,}000 windows per site. Actor tracks and overlapping windows do not cross boundaries; normalization and learned preprocessing use training data only, and test trajectories are not used to construct movement geometry.

\paragraph*{Pooled and held-out-domain evaluation}
The \fmpool{} experiment trains one parameter set over all five domains and evaluates that checkpoint at every site. The \fmzero{} experiment uses leave-one-intersection-out pretraining: for target $q$, the other four intersections form the pretraining set, and the resulting model is applied to $q$ without target-site parameter optimization. Every site serves once as the held-out domain, and results are reported per site before aggregation.

Hyperparameters are selected without the target test interval. Every regime selects the checkpoint with the lowest validation NLL: \fmpool{} uses a balanced all-site validation split, whereas \localmodel{}, \fmzero{}, and \fmfine{} use the target-site validation split. Thus, zero-shot denotes no target parameter optimization; checkpoint selection still uses target validation data. Each learned configuration is represented by one checkpoint, so the comparisons are descriptive and include no confidence intervals or claims of statistical significance.

\section{Pretraining and Deployment Regimes}
\label{sec:training}
The key comparison is not a single-source model transferred to another site. PhaseShift is pretrained over a collection of domains; leave-one-intersection-out evaluation tests whether that multi-domain backbone can be reused at a new site. For $M$ recurring domains, a train-from-scratch workflow can maintain up to $M$ independent parameter sets, whereas PhaseShift seeks one common backbone plus specialization only for the subset of domains where validation supports it.

The four regimes share the same architecture. \localmodel{} is trained from scratch on the target domain; \fmpool{} is one model trained on all known domains; \fmzero{} pretrains on multiple non-target domains and is applied without target parameter optimization; and \fmfine{} initializes from \fmzero{} and is updated using 1{,}000 target windows.

\section{Evaluation}
\label{sec:eval}
\subsection{Conditional trajectory-rollout protocol and baselines}
We evaluate conditional single-vehicle rollouts, not fully interactive traffic simulation. From two seconds of observed history, one target vehicle is rolled autoregressively at 10~Hz for 2, 5, or 10 seconds (up to 100 generated steps). Its generated state is fed back at every step; neighbors, pedestrians, and signal state are replayed from the logged scene. Consequently, the model receives future exogenous context unavailable to a deployed forecaster. Once the generated target diverges from its recorded path, the replayed actors also continue in a world where the target followed the log, so the joint scene can become physically inconsistent. The results therefore quantify long-horizon conditional trajectory generation under recorded context; they do not establish reciprocal multi-agent response, queue propagation, or scene-level traffic simulation.

All learned comparisons use the same partitions, observation history, prediction horizons, and trajectory-sampling budget. We additionally evaluate the supplied constant-acceleration baseline (source setting ``LS-20'') and a calibrated Intelligent Driver Model (IDM)\cite{cal_idm}. These non-learned baselines do not learn a state distribution from logged trajectories, providing a long-horizon check.

\subsection{Metrics}
We report minimum average displacement error (minADE) and minimum final displacement error (minFDE), in meters, over $N{=}6$ sampled rollouts per scenario; lower is better. The 2-, 5-, and 10-s horizons are separate rollouts, not prefixes of one 10-s rollout, each evaluated on 1{,}000 fixed-seed scenarios per site with valid history and future at that horizon. At each step, one of $K{=}25$ mixture modes is drawn and a displacement is sampled from its Gaussian, so the mode can switch between steps. We also report test negative log-likelihood (NLL, in nats) of the ground-truth displacements under the predicted mixture. Test NLL is calculated by feeding the sampled state and averaging over the $N$ samples and the prediction horizon. By contrast, the displacement metrics select the best of six sampled trajectories, an oracle advantage unavailable to a deterministic baseline; those comparisons therefore favor the probabilistic models structurally.

\subsection{One shared model, held-out deployment, and specialization}
Table~\ref{tab:main} reports 2-, 5-, and 10-s results for all five intersections. \cvmodel{} and \idmmodel{} are the supplied classical baselines; \localmodel{} represents five target-specific learned models, \fmpool{} one shared five-domain model, \fmzero{} leave-one-intersection-out pretraining, and \fmfine{} target adaptation of the corresponding zero-shot checkpoint.

\begin{table*}[t]
\centering
\caption{Multi-intersection test results. NLL is in nats; minADE/minFDE are in meters over $N{=}6$ sampled rollouts. Lower is better. Bold marks the best single-run displacement value per site and the best learned-model NLL. Macro averages include all five sites.}
\label{tab:main}
%\scriptsize
\setlength{\tabcolsep}{2.6pt}
\begin{tabular}{llrrrrrrrrr}
\toprule
\shortstack{\textbf{Target}} & \textbf{Method} & \shortstack{\textbf{NLL}\\\textbf{(2 s)}} & \shortstack{\textbf{minADE}\\\textbf{(2 s)}} & \shortstack{\textbf{minFDE}\\\textbf{(2 s)}} & \shortstack{\textbf{NLL}\\\textbf{(5 s)}} & \shortstack{\textbf{minADE}\\\textbf{(5 s)}} & \shortstack{\textbf{minFDE}\\\textbf{(5 s)}} & \shortstack{\textbf{NLL}\\\textbf{(10 s)}} & \shortstack{\textbf{minADE}\\\textbf{(10 s)}} & \shortstack{\textbf{minFDE}\\\textbf{(10 s)}} \\
\midrule
7 & \cvmodel{} & \tph & 1.8559 & 3.6330 & \tph & 2.0404 & 4.2413 & \tph & 2.0216 & 4.3092 \\
7 & \idmmodel{} & \tph & 1.8021 & 3.4662 & \tph & 1.8026 & 3.8486 & \tph & 1.7216 & 4.0604 \\
7 & \localmodel{} & -4.6159 & 0.2800 & 0.5138 & -4.2835 & 0.8262 & 1.8892 & -4.0209 & 2.1595 & 5.1121 \\
7 & \fmzero{} & -4.5001 & 0.4039 & 0.7651 & -5.2363 & 0.7449 & 1.9011 & -5.6398 & 0.9247 & 2.8034 \\
7 & \fmfine{} & -5.0538 & 0.3577 & 0.6388 & \textbf{-5.6513} & 0.6523 & 1.6335 & \textbf{-5.9421} & 0.9263 & 2.9048 \\
7 & \fmpool{} & \textbf{-5.1002} & \textbf{0.2402} & \textbf{0.4105} & -5.3704 & \textbf{0.4781} & \textbf{1.1053} & -5.4502 & \textbf{0.7111} & \textbf{2.0438} \\
\midrule
8 & \cvmodel{} & \tph & 1.3229 & 2.3348 & \tph & 2.1108 & 4.1526 & \tph & 2.5784 & 5.3272 \\
8 & \idmmodel{} & \tph & 1.3854 & 2.4140 & \tph & 1.9943 & 3.8233 & \tph & 2.1031 & 4.6719 \\
8 & \localmodel{} & -4.7131 & 0.3221 & 0.5943 & -4.3360 & 0.7964 & 1.8146 & -4.1724 & 1.8034 & 4.4526 \\
8 & \fmzero{} & -4.6175 & 0.2947 & 0.5667 & -4.4524 & 0.8061 & 1.9511 & -4.6152 & 1.4989 & 3.7417 \\
8 & \fmfine{} & \textbf{-4.9075} & \textbf{0.2673} & \textbf{0.4895} & \textbf{-4.7190} & \textbf{0.6572} & \textbf{1.4897} & \textbf{-4.7516} & \textbf{1.1141} & \textbf{3.0811} \\
8 & \fmpool{} & -4.8790 & 0.2723 & 0.4947 & -4.3897 & 0.7320 & 1.6574 & -4.1268 & 1.6114 & 4.0556 \\
\midrule
9 & \cvmodel{} & \tph & 2.1767 & 4.5921 & \tph & 3.1733 & 7.2096 & \tph & 3.0604 & 8.0948 \\
9 & \idmmodel{} & \tph & 2.1921 & 4.5614 & \tph & 3.0595 & 6.9482 & \tph & 2.7849 & 7.8974 \\
9 & \localmodel{} & -4.1832 & \textbf{0.3308} & \textbf{0.6152} & -3.7845 & \textbf{0.8702} & \textbf{2.0622} & -3.3471 & 2.5617 & 6.4020 \\
9 & \fmzero{} & -3.3586 & 0.4857 & 1.0053 & -3.1008 & 1.0984 & 2.6955 & -2.8843 & 2.9323 & 7.7815 \\
9 & \fmfine{} & \textbf{-4.2184} & 0.3775 & 0.7465 & \textbf{-4.2922} & 0.9959 & 2.5057 & \textbf{-4.5987} & 2.1155 & 5.9349 \\
9 & \fmpool{} & -4.2014 & 0.3375 & 0.6285 & -3.8233 & 0.9124 & 2.2008 & -3.7537 & \textbf{1.9504} & \textbf{5.6599} \\
\midrule
10 & \cvmodel{} & \tph & 1.2499 & 2.5916 & \tph & 1.8278 & 4.0857 & \tph & 2.0966 & 4.5774 \\
10 & \idmmodel{} & \tph & 1.2814 & 2.5997 & \tph & 1.8278 & 3.6114 & \tph & 1.5374 & 3.9786 \\
10 & \localmodel{} & -4.8880 & 0.2538 & 0.4751 & -4.4341 & \textbf{0.5780} & \textbf{1.4760} & -4.1145 & 1.2110 & 3.9279 \\
10 & \fmzero{} & -4.7261 & 0.3047 & 0.5944 & \textbf{-4.8783} & 0.7584 & 2.1138 & \textbf{-5.3643} & 0.8805 & 3.5560 \\
10 & \fmfine{} & \textbf{-5.0065} & 0.2400 & 0.4422 & -4.7273 & 0.6808 & 1.8767 & -4.8208 & 1.1256 & 4.3641 \\
10 & \fmpool{} & -5.0023 & \textbf{0.2087} & \textbf{0.3648} & -4.5303 & 0.5999 & 1.6123 & -4.6125 & \textbf{0.7650} & \textbf{3.0638} \\
\midrule
22 & \cvmodel{} & \tph & 2.0970 & 4.3035 & \tph & 2.4603 & 4.9009 & \tph & 2.5723 & 4.9639 \\
22 & \idmmodel{} & \tph & 2.0424 & 4.0957 & \tph & 1.9869 & 3.9162 & \tph & 1.7738 & 3.9380 \\
22 & \localmodel{} & \textbf{-4.0782} & 0.3302 & 0.5978 & -4.2690 & 0.7339 & 1.6056 & -3.7532 & 1.7956 & 4.5589 \\
22 & \fmzero{} & -3.3864 & 0.4492 & 0.8483 & -4.2868 & 0.7399 & 1.7330 & -4.6380 & 1.1098 & 3.2940 \\
22 & \fmfine{} & -4.0660 & 0.3523 & 0.6506 & \textbf{-4.8381} & 0.5703 & 1.3018 & \textbf{-5.0950} & \textbf{0.9211} & 2.8542 \\
22 & \fmpool{} & -3.9896 & \textbf{0.3068} & \textbf{0.5378} & -4.3143 & \textbf{0.5398} & \textbf{1.1887} & -3.9333 & 0.9603 & \textbf{2.6862} \\
\midrule
Macro avg. & \cvmodel{} & \tph & 1.7405 & 3.4910 & \tph & 2.3225 & 4.9180 & \tph & 2.4659 & 5.4545 \\
Macro avg. & \idmmodel{} & \tph & 1.7407 & 3.4274 & \tph & 2.1342 & 4.4295 & \tph & 1.9842 & 4.9093 \\
Macro avg. & \localmodel{} & -4.4957 & 0.3034 & 0.5592 & -4.2214 & 0.7609 & 1.7695 & -3.8816 & 1.9062 & 4.8907 \\
Macro avg. & \fmzero{} & -4.1177 & 0.3876 & 0.7560 & -4.3909 & 0.8295 & 2.0789 & -4.6283 & 1.4692 & 4.2353 \\
Macro avg. & \fmfine{} & \textbf{-4.6504} & 0.3190 & 0.5935 & \textbf{-4.8456} & 0.7113 & 1.7615 & \textbf{-5.0416} & 1.2405 & 3.8278 \\
Macro avg. & \fmpool{} & -4.6345 & \textbf{0.2731} & \textbf{0.4873} & -4.4856 & \textbf{0.6524} & \textbf{1.5529} & -4.3753 & \textbf{1.1996} & \textbf{3.5019} \\
\bottomrule
\end{tabular}
\end{table*}

\begin{figure}[t]
\centering
\includegraphics[width=\columnwidth]{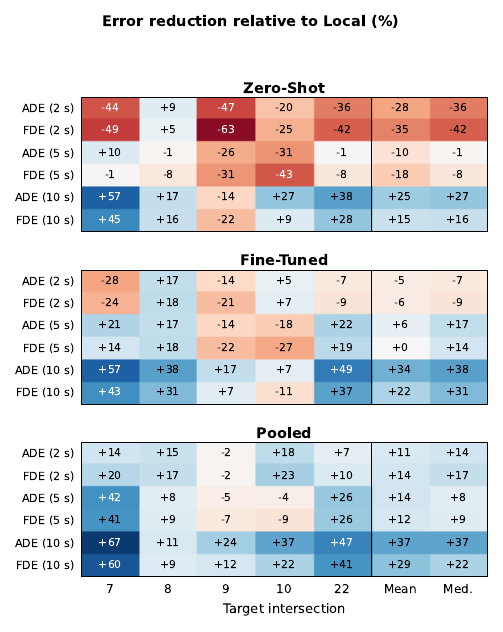}
\caption{Domain-level error reduction relative to independently trained \localmodel{} models. Positive values mean the shared-backbone regime has lower error; negative values indicate a residual specialization gap. ``Mean'' and ``Med.'' summarize site-level percentage reductions.}
\Description{Three vertically stacked diverging heatmaps compare Zero-Shot, Fine-Tuned, and Pooled performance with Local models across sites 7, 8, 9, 10, and 22. All three regimes show strong 10-second gains at most sites, short-horizon losses for Zero-Shot, and the pooled model is strongest on the aggregate.}
\label{fig:relative-results}
\end{figure}

Figure~\ref{fig:relative-results} expresses each shared-backbone regime relative to \localmodel{}; negative cells indicate a residual specialization gap.

\paragraph*{Pooled consolidation}
The central test is whether harmonized data can support one common model instead of one parameter set per intersection. At 10~s, \fmpool{} improves both metrics over \localmodel{} at all five sites, with median site-level reductions of 36.8\%/22.0\% and ratio-of-macro-average reductions of 37.1\%/28.4\%. The largest gains occur at site~7 (67.1\%/60.0\%) and the short-duration site~22 (46.5\%/41.1\%). At 2~s, \fmpool{} improves both metrics at four sites and trails \localmodel{} only at site~9, by about 2\%; at 5~s it improves both metrics at sites 7, 8, and 22 while conceding small margins at sites 9 and 10. Thus, one shared checkpoint remains competitive at short horizons and has its clearest advantage at 10~s. Because \fmpool{} sees five times the total data of each \localmodel{}, the fixed-volume sweep below provides a first control at site~7.

\paragraph*{Unseen-domain deployment}
Each \fmzero{} model is pretrained on four source domains and applied to a new intersection without target-site parameter optimization. At 10~s, it beats the target's own \localmodel{} on both metrics at four of five sites; only site~9 degrades ($-$14.5\%/$-$21.5\%). In the single cross-region fold at site~22, a backbone trained on four Gainesville sites reduces the South Florida target's 10-s errors by 38.2\%/27.7\%. One fold cannot establish broad cross-region generalization. The short-horizon pattern reverses: at 2~s, \fmzero{} trails \localmodel{} everywhere except site~8, with median changes of $-$36.0\%/$-$41.9\%. These comparisons associate target data with short-horizon precision and multi-source pretraining with lower long-horizon error. They do not identify the mechanism, because \fmzero{} also sees four times the training volume of \localmodel{}.

\paragraph*{Specialization signal}
The zero-shot gap is an operational signal, not an attribution to topology. Target validation can indicate whether to retain \fmzero{}, apply \fmfine{}, or inspect the site as a distinct domain.

\subsection{Target fine-tuning across intersections}
\fmfine{} updates each zero-shot checkpoint with 1{,}000 target windows, 1\% of the 100{,}000-window \localmodel{} training budget; checkpoint selection uses the separate target validation split in Section~\ref{sec:data}. At 10~s, fine-tuning lowers both zero-shot errors at sites~8, 9, and 22, leaves site~7 essentially unchanged (0.9247$\to$0.9263 minADE), and degrades site~10 (0.8805$\to$1.1256). Relative to \localmodel{}, it improves minADE at all five sites and minFDE at four, with median reductions of 38.2\% and 30.8\%; site~10 minFDE is the exception ($-$11.1\%). \fmfine{} is the best regime at site~8 on every horizon and metric and attains the best 10-s minADE at site~22. It is not universally beneficial: it trails \localmodel{} at sites~7, 9, and 22 at 2~s and at sites~9 and 10 at 5~s. The adaptation follows roughly 400{,}000 source windows and uses target-derived geometry, so its total information exceeds the local model's. Without a matched 1{,}000-window scratch model, these results establish the performance of the adaptation recipe, not the isolated value of pretrained initialization. Fine-tuning should therefore be selected on validation at the intended horizon.

\subsection{Fixed-volume composition sweep}
To separate composition from nominal sample count, we fix the site-7 training budget at 100{,}000 windows and vary only the site mixture (Table~\ref{tab:sweep}), using the nested sampling in Section~\ref{sec:data}; 100/0/0/0/0 is the site-7 \localmodel{}. Every cross-site mixture improves every horizon over pure-local training. Replacing 20\% of the local windows with equal shares from the other four sites cuts 10-s error by 66.2\%/58.4\%. The local-heavy 80/5/5/5/5 mixture is best at 10~s and comes within 2.8\%/4.0\% of \fmpool{}, despite using one fifth as many total windows. The decisive contrast is therefore between pure-local and mixed-site training, not among the mixtures: at this target, a modest cross-site injection is sufficient, and more non-local data is not uniformly better.

Two boundaries remain. The sweep fixes window count, not independent temporal evidence: site~7's local windows cover roughly nineteen signal cycles, whereas the mixtures draw from many more cycles across five recordings. Best-of-$N$ scoring can also reward wider output dispersion, although test NLL improves sharply at 10~s ($-5.97$ for 80/5/5/5/5 and $-5.82$ for 60/10/10/10/10, versus $-4.02$ for pure-local), arguing against a dispersion-only explanation. Same-site temporal controls, single-source mixtures, paired seeds, and other targets are needed before generalizing.

\begin{table}[t]
\centering
\caption{Fixed-100{,}000-window composition sweep at site~7. Splits give the percentage drawn from sites 7/8/9/10/22; 100/0/0/0/0 is \localmodel{}. Bold marks the best value per column. One run per composition.}
\label{tab:sweep}
\scriptsize
\setlength{\tabcolsep}{2.2pt}
\begin{tabular}{lrrrrrr}
\toprule
\shortstack[l]{\textbf{Split}\\\textbf{7/8/9/10/22}} & \shortstack{\textbf{minADE}\\\textbf{(2 s)}} & \shortstack{\textbf{minFDE}\\\textbf{(2 s)}} & \shortstack{\textbf{minADE}\\\textbf{(5 s)}} & \shortstack{\textbf{minFDE}\\\textbf{(5 s)}} & \shortstack{\textbf{minADE}\\\textbf{(10 s)}} & \shortstack{\textbf{minFDE}\\\textbf{(10 s)}} \\
\midrule
100/0/0/0/0    & 0.2800 & 0.5138 & 0.8262 & 1.8892 & 2.1595 & 5.1121 \\
80/5/5/5/5     & 0.2492 & 0.4268 & 0.4774 & 1.2210 & \textbf{0.7309} & \textbf{2.1264} \\
60/10/10/10/10 & \textbf{0.2241} & \textbf{0.3843} & 0.4912 & 1.2340 & 0.8315 & 2.5654 \\
40/15/15/15/15 & 0.2599 & 0.4607 & 0.5560 & 1.2933 & 1.0697 & 2.9718 \\
20/20/20/20/20 & 0.2438 & 0.4015 & \textbf{0.4641} & \textbf{1.0681} & 0.9986 & 2.8753 \\
\bottomrule
\end{tabular}
\end{table}

\subsection{Classical baselines and long-horizon behavior}
Every learned regime has substantially lower best-of-sample error than \cvmodel{} and \idmmodel{} at 2 and 5~s at every site, subject to the unequal hypothesis budget noted above. At 10~s, however, \localmodel{} falls behind both classical baselines at site~7 (2.1595/5.1121, versus 1.7216/4.0604 for IDM and 2.0216/4.3092 for constant acceletaion) and behind IDM at site~22 (1.7956/4.5589, versus 1.7738/3.9380). These are the two highest-flow sites and the only sites with this reversal. The cross-domain regimes avoid it in our runs: \fmpool{} reaches 0.7111/2.0438 at site~7, 58.7\%/49.7\% below IDM, and every pretrained-backbone regime remains well ahead of IDM at site~22. At the other sites, the learned regimes also remain ahead of both classical baselines at 10~s, apart from two single-metric exceptions: site-9 \fmzero{} minADE and site-10 \fmfine{} minFDE.

The reversal is consistent with covariate shift under sequential prediction: autoregressive rollout feeds generated states back into later contexts, allowing errors to compound as the model moves away from the logged-state distribution~\cite{ross2011dagger,zhang2025catk}. The design makes that exposure explicit: training supervises 20 future steps (2~s), whereas the 10-s evaluation rolls out 100 steps. The experiment establishes the reversal and its absence in the cross-domain regimes; it does not isolate covariate shift as the cause.

\paragraph*{Likelihood}
NLL provides a selection-free check on the sampled displacement metrics. At 10~s, a pretrained-backbone regime has the best NLL at every site: \fmfine{} at sites 7, 8, 9, and 22, and \fmzero{} at site~10. The margin over \localmodel{} can be large ($-5.94$ versus $-4.02$ at site~7), showing that the long-horizon gains are not solely an artifact of wider output dispersion under best-of-$N$ scoring. At 2~s the pattern is mixed, including the best NLL for \localmodel{} at site~22, consistent with the less uniform short-horizon displacement results.

\subsection{Per-site synthesis}
The site-level results clarify where consolidation helps most, although the observational comparisons do not identify why the sites differ.

Site~7 is the busiest Gainesville domain (16 movement classes, roughly 2673 vehicles per hour, and a 150-s cycle) and shows the sharpest long-horizon failure. \localmodel{} rises from $0.28$ minADE at 2~s to $2.16$ at 10~s and falls behind calibrated IDM. Every cross-domain regime has much lower 10-s error; \fmpool{} is best in all six site-7 displacement cells and cuts the 10-s errors by 67.1\%/60.0\%. The fixed-volume sweep strengthens the composition result: every tested cross-site mixture removes most of the degradation. Repeated seeds and rollout-drift diagnostics are still needed to explain the mechanism.

Site~22 demonstrates the complementary value of consolidation when local temporal coverage is narrow. Its 100{,}000 windows come from only 19 minutes, roughly six signal cycles, while the site also has the highest demand (roughly 5062 vehicles per hour), the largest footprint (24 movement classes), and a different region from every source domain. \localmodel{} reaches $1.80$/$4.56$ at 10~s; \fmzero{} trained on the four Gainesville sites improves this to $1.11$/$3.29$, and \fmpool{} reaches $0.96$/$2.69$. In this single cross-region case, \fmfine{} and \fmpool{} split the best 10-s metrics and both decisively outperform local training.

Site~9 is the clearest specialization signal. It is the smallest, lowest-flow intersection (8 movement classes and roughly 985 vehicles per hour) and has the most atypical signal timing (46\% mean green split, versus 20--28\% elsewhere). It is the only site where \fmzero{} degrades both 10-s metrics ($-$14.5\%/$-$21.5\%) and where \fmpool{} concedes short-horizon accuracy to \localmodel{}. Yet \fmpool{} is still the best 10-s regime ($1.95$ versus $2.56$ minADE), marking site~9 as the domain least covered at short horizons rather than a failure of pooled consolidation.

Sites~8 and 10 occupy the middle ground. At site~8, \fmfine{} is best on every horizon and metric. At site~10, which has the longest training span (84 minutes), \fmpool{} still wins at 2 and 10~s, while 10-s fine-tuning is counterproductive ($0.88\to1.13$ minADE versus \fmzero{}). Across sites, the same backbone therefore provides large long-horizon gains where local models degrade (7 and 22), consolidation where local data are stronger (8 and 10), and a clear signal for horizon-specific specialization (9).

\section{Discussion}
Three findings define the result. First, one pooled checkpoint outperforms the five independently trained local models in aggregate at every horizon and on both displacement metrics, with gains at every site at 10~s. Second, leave-one-site-out reuse is strongly horizon-dependent: it improves both 10-s metrics at four sites but usually sacrifices 2-s accuracy. Third, at site~7, every fixed-volume cross-site mixture sharply outperforms pure-local training, so the pooled gain is not explained by nominal window count alone. Together, these results motivate a practical lifecycle: consolidate known domains, measure coverage at a new domain, and specialize only when validation shows a meaningful residual. The experiments compare all candidate regimes, but they do not execute this validation-driven selection policy end to end.

\paragraph*{Operational consolidation}
At 10~s, \fmpool{} is the best learned regime at sites~7, 9, and 10 and is within 5\% of the per-metric best at site~22; site~8 is the only site where \fmfine{} wins both metrics. Retrospectively, the five-site portfolio therefore needs only two parameter sets: \fmpool{} plus the 1{,}000-window \fmfine{} specialist for site~8. Because this assignment is chosen from test results, it illustrates consolidation potential rather than a validated deployment policy; operational selection must use validation under one fixed objective.

\paragraph*{Physical-AI perspective}
Because physical layout, control, demand, road-user behavior, and sensing jointly shape the observations, harmonization is an interface between learned models and infrastructure rather than a preprocessing convenience. PhaseShift shows that one representation can bridge five realized physical--control regimes while preserving useful site context. The experiments do not separate the contributing factors or establish robustness to any one of them; that requires stratification by movement type, signal phase, queue state, pedestrian presence, and track completeness.

The same lifecycle can treat recurring demand, signal-plan, school-period, or approach-imbalance conditions as site--condition domains. Persistent residuals identify where to investigate; attributing them to topology or another factor still requires metadata and matched comparisons.

\section{Limitations}
\paragraph*{Evidence and data scope}
Each learned configuration is represented by one checkpoint, without uncertainty estimates; rankings separated by a few percent may be seed-dependent. The balanced training sets span only 19--84 minutes per site and therefore provide limited independent temporal evidence. \fmpool{} sees five times the data of each \localmodel{}, and each \fmzero{} model sees four times as much. The composition sweep controls nominal window count, but not independent temporal coverage or the potential benefit of wider output dispersion under best-of-$N$ scoring. It is also limited to one target, and no ablation isolates the harmonization representation.

\paragraph*{Evaluation boundary}
Best-of-six displacement metrics structurally favor the probabilistic models over the single-output deterministic baselines, and no contemporary learned baseline is evaluated. NLL avoids sample selection but has no deterministic counterpart. The protocol measures conditional generation under replayed exogenous context, not forecasting or closed-loop simulation; top-1 or expected error and calibration are not reported. Checkpoint selection for every regime, including \fmzero{}, uses the target site's validation NLL.

\paragraph*{Attribution and adaptation}
The fine-tuning comparison lacks a matched 1{,}000-window scratch control, so it does not isolate the value of pretrained initialization. Physical, control, demand, human, and sensing factors co-vary in the recordings; their causal effects are not separated, and results are not stratified by movement class, signal phase, or observation completeness. Finally, the IDM crossover at site~7 shows that short-horizon fit does not guarantee accurate long autoregressive rollout. The system includes no on-policy or closed-loop mechanism to address covariate shift.

\section{Conclusion}
PhaseShift replaces a train-from-scratch default with one topology-aware representation and a reusable multi-intersection backbone. In this five-site study, the pooled model achieves the best macro displacement error at every horizon and improves both 10-s metrics over local training at every site. Zero-shot reuse, including one cross-region fold, wins both 10-s metrics at four sites; 1{,}000-window fine-tuning is best at one. The pretrained regimes also avoid the long-horizon reversal observed when local models fall behind calibrated IDM at the two highest-flow sites. At site~7, every cross-site mixture produces large gains at a fixed nominal window count, demonstrating that composition matters in addition to count for that target. The remaining evidence gap is attribution: repeated seeds, matched temporal controls, additional targets, and representation ablations are needed to determine why cross-site training works. Within the evaluated replay-conditioned setting, PhaseShift demonstrates that real intersection data can be consolidated into a stronger shared long-horizon model while retaining specialization where the operating domain requires it.

\section*{Reproducibility Statement}
An accompanying artifact documents the data partitions, preprocessing and geometry procedures, model and training configuration, and evaluation settings for the reported comparisons.

% References begin in the unused portion of the final content page.
\IEEEtriggeratref{18}
\bibliographystyle{ieeetr}
\bibliography{references}

\clearpage

\appendices
\section{Artifacts}
\subsection{Data Preprocessing}
Data is partitioned in a ratio of 80/10/10 Train/Test/Validate based on total time of the collected data. Each of the split is preprocessed separately and trajectories broken by this split removed from the scene in order to reduce the risk of data-leakage. Preprocessing is required to weed out noisy and incomplete trajectories from complete trajectories. We use starting and ending bounding box to mark trajectories belonging to specific movement class. These trajectories are used to create movement polylines. We only require a single trajectory belonging to movement class to generate the movement polylines. An analysis on the number of trajectories required to create the polylines with its impact on the downstream metrics has not been done in this study. To find the stop line, we draw a line half vehicle distance away from the center point of the stopped vehicle at the intersection. Ideally, even at an intersection with low traffic volume, a couple of cycle lengths worth of data is enough to generate a good movement polylines and stop lines.

\subsection{Train/Fine-tune}
The complete list of hyperparameters used in training and finetuning is mentioned in the table \ref{tab:hyperparams} and \ref{tab:finetune_hyperparams}. We use the same architecture and hyperparameters while training and finetuning across the five sites. For fine-tuning a 1{,}000 size subset is sampled from the 100{,}000 size training split. Finetune updates all the weights across the model. We have not analyzed the impact of freezing weights of different components of the model and its impact on the evaluation as part of this paper.

\begin{table}[t]
\centering
\caption{Model and training hyperparameters.}
\label{tab:hyperparams}
\begin{tabular}{ll}
\toprule
\textbf{Parameter} & \textbf{Value} \\
\midrule
\multicolumn{2}{l}{\textit{Scene context}} \\
Time step $\Delta t$                    & \SI{0.1}{\second} \\
Observation history $T_h$               & 20 steps (\SI{2.0}{\second}) \\
Prediction horizon $T_f$                & 20 steps (\SI{2.0}{\second}) \\
Max.\ neighbors                         & 10 \\
Neighbor radius                         & \SI{15}{\meter} \\
Max.\ map polylines                     & 10 \\
Vectors per polyline                    & 3 \\
Map / signal attention radius           & \SI{15}{\meter} \\
Speed normalization factor              & \SI{20}{\meter\per\second} \\
\midrule
\multicolumn{2}{l}{\textit{Architecture}} \\
Embedding dimension $d_{\text{model}}$  & 256 \\
Spatial attention layers                & 8 \\
Spatial attention heads                 & 4 \\
Temporal decoder layers                 & 8 \\
Temporal decoder heads                  & 4 \\
Dropout                                 & 0.1 \\
\midrule
\multicolumn{2}{l}{\textit{Output distribution}} \\
Output parameterization                 & Cartesian GMM \\
Mixture components                      & 25 \\
Training objective                      & GMM negative log-likelihood \\
\midrule
\multicolumn{2}{l}{\textit{Optimization}} \\
Optimizer                               & Adam \\
Base learning rate                      & $1\times10^{-3}$ \\
LR schedule                             & Transformer warmup--decay \\
Warmup steps                            & 4000 \\
Batch size                              & 256 \\
Epochs                                  & 100 \\
Early-stopping patience                 & 10 epochs \\
\bottomrule
\end{tabular}
\end{table}

\begin{table}[t]
\centering
\caption{Fine-tuning hyperparameters. All values not listed are inherited
unchanged from the pre-trained model (Table~\ref{tab:hyperparams}).}
\label{tab:finetune_hyperparams}
\begin{tabular}{lcc}
\toprule
\textbf{Parameter} & \textbf{Pre-training} & \textbf{Fine-tuning} \\
\midrule
Learning rate               & $1\times10^{-3}$ & $1\times10^{-5}$ \\
Batch size                  & 256              & 64 \\
Warmup steps                & 4000             & 20 \\
\midrule
Training windows $M$        & 100{,}000         & 1{,}000 \\
\bottomrule
\end{tabular}
\end{table}

\subsection{Least-Squares and IDM Baselines}
\label{app:baselines}

We compare against two non-learned longitudinal predictors, both rolled out for the same $H$ steps from the ego's ground-truth state at the last observed frame. The \emph{least-squares} baseline fits a first-order polynomial to the ego's speed over the history window by ordinary least squares and extrapolates it forward, i.e.\ a constant-acceleration model whose acceleration is the fitted slope. The \emph{IDM} baseline instead sets the acceleration at each step from the Intelligent Driver Model,
\begin{align}
\dot{v} &= \alpha\left[1 - \left(\frac{v}{v_0}\right)^{\delta} - \left(\frac{s^{*}(v,\Delta v)}{s}\right)^{2}\right], \\
s^{*}(v,\Delta v) &= s_0 + \max\!\left(0,\; vT + \frac{v\,\Delta v}{2\sqrt{\alpha\beta}}\right),
\end{align}
with $\delta = 4$, where $s$ is the bumper-to-bumper gap to the binding leader and $\Delta v$ the approach rate. The leader is the nearer of (i) the closest vehicle inside a \ang{30} forward cone around the ego heading and (ii) a virtual stationary leader placed at the cluster stop line whenever the ego's signal phase is not green; if neither exists the ego is in free flow and the interaction term vanishes. Parameters $(v_0, s_0, T, \alpha, \beta)$ are set to the posterior means of a Bayesian calibration \cite{cal_idm} performed per intersection on the training split.

At every step the predicted position is projected onto the nearest point of the ego's movement polyline and the lane tangent becomes the new heading, so both baselines are constrained to the map in the same way.

\end{document}